\documentclass[journal]{IEEEtran}
\ifCLASSINFOpdf
\else
\fi
\begin{document}
\title{Diffusion Models in NLP: A Survey}
\author{
Yuansong Zhu,
        Yu Zhao,~\IEEEmembership{Member,~IEEE}
        % <-this % stops a space

\IEEEcompsocitemizethanks{
\IEEEcompsocthanksitem Y. Zhu and Y. Zhao
 % ,Q. Li and X. Chen are 
is with Fintech Innovation Center, Financial Intelligence and Financial Engineering Key Laboratory of Sichuan Province, Institute of Digital Economy and Interdisciplinary Science Innovation, Southwestern University of Finance and Economics, China. \protect

% \IEEEcompsocthanksitem H. Du
% , H. Du and G. Kou are 
% is with School of Business Administration, Southwestern University of Finance and Economics,  China.\protect
% E-mail: dhmfcc@smail.swufe.edu.cn
}
}

% \markboth{Journal of \LaTeX\ Class Files,~Vol.~14, No.~8, August~2015}%
% {Shell \MakeLowercase{\textit{et al.}}: Bare Demo of IEEEtran.cls for IEEE Journals}



\maketitle

\begin{abstract}
Diffusion models have become a powerful family of deep generative models, with record-breaking performance in many applications. This paper first gives an overview and derivation of the basic theory of diffusion models, then reviews the research results of diffusion models in the field of natural language processing, from text generation, text-driven image generation and other four aspects, and analyzes and summarizes the relevant literature materials sorted out, and finally records the experience and feelings of this topic literature review research.
\end{abstract}


% \fi
\IEEEpeerreviewmaketitle

\section{Introduction}
Generative models are a class of models that can randomly generate observation results based on some hidden parameters, and they are usually used for natural language processing, image generation, audio generation and other tasks. Among them, VAE and GAN are well-known generative models. In recent years, with the further development of generative models, diffusion models have become a new type of generative model with their powerful generation ability. Nowadays, they have achieved great achievements and have been widely used in computer vision, speech generation, natural language processing and other aspects

With the proposal of diffusion model DDPM in 2020, it has attracted wide attention in the field of artificial intelligence. In the following 2021 and 2022, diffusion models began to shine, and the number of related literature publications showed a blowout growth. After achieving great success in the initial field of image generation, diffusion models have now also entered the field of natural language processing and achieved some results. In order to summarize and summarize the research results of diffusion models in the field of natural language processing, this paper has done the following work:
\subsubsection{}Based on the original DDPM model, we introduce the basic principles and formulas of the diffusion model.

\subsubsection{}Organize and review the research results of diffusion models in the field of NLP.

\subsubsection{}Information such as code and citation volume of cited literature is collected and analyzed.
% for IEEE journal papers produced under \LaTeX\ using
% IEEEtran.cls version 1.8b and later.
% I wish you the best of success.

 

% \subsection{Subsection Heading Here}
% Subsection text here.

%\IEEEpubidadjcol

% \subsubsection{Subsubsection Heading Here}
% Subsubsection text here.

\section{Preliminaries}
Diffusion models are generative models, which means they are used to generate data similar to the data they are trained on. Essentially, diffusion models work by continuously adding Gaussian noise to corrupt the training data, and then learning to recover the data by reversing this noise process. After training, we can use the diffusion model to generate data by simply passing random samples of noise through the learned denoising process.

The diffusion model consists mainly of a diffusion process and an inverse diffusion process, the idea of which is derived from nonequilibrium thermodynamics. In the diffusion process a parametric Markov chain is defined, starting from the initial data distribution and adding Gaussian noise to the data distribution at each step for a duration of times. With a suitable setting, the original data gradually loses his characteristics as it continues to increase in size. We can understand that after an infinite number of noise addition steps, the final data becomes a picture with no features and completely random noise, the "snow screen" we talked about at the beginning.

The inverse diffusion process starts with a random noise and is gradually reduced to the original image without noise, i.e. the denoising process, the inverse diffusion process is in fact the process of generating the data. It is important to note that the inverse process still ends up generating an image that is related to the original data. If you input a picture of a cat, the image generated by the model should be a cat, and if you input a picture of a dog, the image generated should also be related to the dog.

\subsection{Diffusion Process}
The diffusion process is the process of adding noise to a picture, which is free of learnable parameters, and $X_{t+1}$ is obtained by adding noise to $X_{t}$, which is only affected by $X_{t}$. The diffusion process is therefore a Markov process, and the step size of each diffusion step is affected by the variation  $ {\beta _ {t}\in (0,1)}^ {T}_ {t=1} $ 
 and $ \beta _ {1} < \beta _ {2} < \cdots < \beta _ {T} $ ,
, which indicates that the noise added is increasing. $q$( $ X_ {t} $ $ |X_ {t-1} $ )
 can be written in the following form, given the condition $X_{t-1}$, obeying a normal distribution with mean ( $ X_ {t} $ $ |X_ {t-1} $ )
 and variance ( $ X_ {t} $ $ |X_ {t-1} $ )
:
\begin{equation}
q({X_t}|{X_{t - 1}}) = N({X_t};\sqrt {1 - {\beta _t}}  \cdot {X_{t - 1}};{\beta _t}{\rm I})
\end{equation}

Using the reparameterization technique to represent $X_{t}$,such that 
${a_t} = 1 - {\beta _t}$,
${Z_t} \sim {\rm N}(0,1),{\rm{t}} \ge 0$, then:
\begin{equation}
{X_t} = \sqrt {{a_t}} {X_{t - 1}} + \sqrt {1 - {a_t}} {Z_{t - 1}}
\end{equation} 
In calculation ${\rm{q}}({X_t}|{X_0})$, since it is defined as a Markov chain, the joint probability distribution of ${X_{1:{\rm{T}}}}$ given ${X_0}$ is :
\begin{equation}
    q({X_{1:{\rm{T}}}}|{X_0}) = \mathop \prod \limits_{t = 1}^{\rm{T}} q({X_t}|{X_{t - 1}})
\end{equation}
The calculation of ${\rm{q}}({X_t}|{X_0})$ requires constant iteration, and to simplify the operation, it is desirable that given ${X_0}$,${\beta _t}$ it be calculated, and given ${a_t} = 1 - {\beta _t}$,$\mathop a\limits^\_  = \prod _{t = 1}^T{a_t}$, we have:
\begin{equation}
    \begin{array}{c}
{X_t} = \sqrt {{a_t}} {X_{t - 1}} + \sqrt {1 - {a_t}} {Z_{t - 1}}\\
 = \sqrt {{a_t}} (\sqrt {{a_{t - 1}}} {X_{t - 2}} + \sqrt {1 - {a_{t - 1}}} {Z_{t - 2}}) + \sqrt {1 - {a_t}} {Z_{t - 1}}\\
 = \sqrt {{a_t}{a_{t - 1}}} {X_{t - 2}} + \sqrt {{a_t} - {a_t}{a_{t - 1}}} {Z_{t - 2}} + \sqrt {1 - {a_t}} {Z_{t - 1}}\\
 = \sqrt {{a_t}{a_{t - 1}}} {X_{t - 2}} + \sqrt {1 - {a_t}{a_{t - 1}}} {\overline Z _{t - 2}}\\
 =  \ldots \\
 = \sqrt {{{\overline a }_t}} {X_0} + \sqrt {1 - {{\overline a }_t}} \overline Z 
\end{array}
\end{equation}
In summary, the derivation of the diffusion process with respect to A is:
\begin{equation}
    {X_t} = \sqrt {{{\overline a }_t}} {X_0} + \sqrt {1 - {{\overline a }_t}} \overline Z ,\overline Z  \sim N(0,{\rm I})
\end{equation}
\begin{equation}
{\rm{q}}({X_t}|{X_0}) = N({X_t};\sqrt {{a_t}} {X_0};(1 - {\overline a _t}{\rm I}))
\end{equation}
\subsection{Inverse Diffusion Processes}
If the distribution of ${X_{t - 1}}$ can be known under the given condition ${X_t}$, that is, if ${\rm{q}}({X_{t - 1}}|{X_t})$ can be known, then a picture can be obtained from any noise picture by sampling again and again to achieve the purpose of picture generation, but it is obviously difficult to know ${\rm{q}}({X_{t - 1}}|{X_t})$, so it needs to be used to approximate ${p_\theta }({X_{t - 1}}|{X_t})$. ${p_\theta }({X_{t - 1}}|{X_t})$ is the network that needs to be trained. Since the noise added each time during the diffusion process is very small, it is assumed that ${p_\theta }({X_{t - 1}}|{X_t})$ is also a Gaussian distribution, which can be fitted by a neural network. The inverse process is also a Marko husband chain process.
\begin{equation}
    {p_\theta }({X_{t - 1}}|{X_t}) = N({X_{t - 1}};{\mu _\theta }({X_t},t),\sum\limits_\theta  {({X_t},t)} )
\end{equation}
\begin{equation}
    {p_\theta }({X_{0:{\rm{T}}}}) = P({X_{\rm{T}}})\mathop \prod \limits_{t = 1}^{\rm{T}} {p_\theta }({X_{t - 1}}|{X_t})
\end{equation}
Although ${\rm{q}}({X_{t - 1}}|{X_t})$ is not known, ${\rm{q}}({X_{t - 1}}|{X_t}{X_0})$ can be represented by ${\rm{q}}({X_t}|{X_0})$ and ${\rm{q}}({X_t}|{X_{t - 1}})$, i.e., ${\rm{q}}({X_{t - 1}}|{X_t}{X_0})$ is knowable, and the following derivation is made for ${\rm{q}}({X_{t - 1}}|{X_t}{X_0})$.
\begin{equation}
    \begin{array}{c}
{\rm{q}}({X_{t - 1}}{X_t}{X_0}) = \frac{{q({X_{t - 1}}{X_t}{X_0})}}{{q({X_t}{X_0})}}\\
 = \frac{{q({X_{t - 1}}{X_t}{X_0})}}{{q({X_{t - 1}}{X_0})}}\frac{{q({X_{t - 1}}{X_0})}}{{q({X_t}{X_0})}}\\
 = {\rm{q}}({X_t}{X_{t - 1}}{X_0}) \cdot \frac{{q({X_{t - 1}}{X_0})}}{{q({X_t}{X_0})}}
\end{array}
\end{equation}
Since the diffusion process is a Markov process, the next step :
\begin{equation}
    {\rm{q}}({X_t}{X_{t - 1}}{X_0}) = q({X_t}{X_{t - 1}})
\end{equation}
\begin{equation}
    {\rm{q}}({X_{t - 1}}{X_t}{X_0}) = q({X_t}{X_{t - 1}}) \cdot \frac{{q({X_{t - 1}}{X_0})}}{{q({X_t}{X_0})}}
\end{equation}
So far, ${\rm{q}}({X_{t - 1}}{X_t}{X_0})$ has been expressed in terms of $q({X_t}{X_0})$ and $q({X_t}{X_{t - 1}})$. The expression for ${\rm{q}}({X_{t - 1}}{X_t}{X_0})$ is as follows:
\begin{equation}
    {\rm{q}}({X_{t - 1}}{X_t}{X_0}) = N({X_{t - 1}};\frac{1}{{\sqrt {{a_t}} }}({X_t} - \frac{{{\beta _t}}}{{\sqrt {1 - {{\overline a }_t}} }}Z);\frac{{1 - {{\overline a }_{t - 1}}}}{{1 - {{\overline a }_t}}}{\beta _t}),
\end{equation}
$Z \sim N(0,{\rm I})$
\section{Application in NLP}
Research on diffusion models in the field of natural language processing started in 2021, and a large number of results emerged in 2022. So far in the field of NLP, the diffusion model relies on its own powerful generation ability and has achieved more results mainly in text generation and text-driven image generation, and has also made some progress in the research of text-to-speech, This section reviews and organizes the literature on the diffusion model from four aspects: text generation, text-driven image generation, text-to-speech, and others, and outlines its application results in the NLP field. This section reviews and organizes the literature on diffusion models from four aspects: text generation, text-driven image generation, text-to-speech, and others, and outlines the results of their applications in NLP.
\subsection{Text Generation}
Natural language processing aims to understand, model and manage human language from different sources (e.g., text or audio). Text generation has become one of the most critical and challenging tasks in natural language processing. Its goal is to write seemingly reasonable and readable text in human language given input data (e.g., sequences and keywords) or random noise. Many methods based on diffusion models have been developed for text generation.

Ting Chen\cite{chen2022analog} proposes a diffusion model for generating discrete data and applies it to the image captioning task. The main idea behind their approach is to first represent discrete data as binary bits and then train a continuous diffusion model that models these bits as real numbers, called simulated bits.

The natural language processing research group at Stanford University, Xiang Lisa Li \cite{li2022diffusion} et al. exploit the non-autoregressive mechanism of the diffusion model to accomplish the task of controlled text generation by iteratively noise reduction from noise vectors to word vectors with the help of the continuous diffusion property, which is innovative in defining an equation of word embedding unifying the discrete to continuous states in the diffusion process.

Jacob Austin \cite{austin2021structured} team proposed the diffusion-like generative model D3PM for discrete data, which improves the diffusion model for discrete data by defining a new discrete damage process and generalizes the diffusion mode in the diffusion process.

Shansan Gong \cite{gong2022diffuseq} team proposed a diffusion model for sequence-to-sequence (SEQ2SEQ) text generation task design using end-to-end classifier-free guided diffusion condition generation to overcome the problem of discrete nature of text and provided provided a comparative link between the diffusion model and the autoregressive and non-autoregressive models.

Guangyi Liu's \cite{liu2022composable} team proposed a new method for composable text manipulation in a compact text latent space. The diffusion process is put in charge of the control of text properties on a low-dimensional continuous space only. And after making the latent vectors have the corresponding properties, the latent vectors are then handed over to the decoder to generate the text. The limitation of fixed-length generation of Diffusion-LM is avoided, and the text fluency and model training speed are improved.

Zhengfu He's \cite{he2022diffusionbert} team (Xipeng Qiu's group) used diffusion models to improve the generative masked language model DiffusionBERT, which solves the unconditional text generation problem of non-autoregressive models combining pre-trained language models with discrete diffusion models of the absorption state of the text, achieving significant improvements over the previous text diffusion models (e.g., D3PM and diffusion - lm) and the previous generative masked language models.

Nikolay Savinov's \cite{savinov2021step} team proposed a new text generation model, Stepwise Unfolding Noise Reduction Self-Encoder (SUDAE), which does not rely on an autoregressive model and converges using fewer iterations than usual diffusion methods, while qualitatively producing better samples on natural language datasets, achieving better results in English-German translation tasks.
\subsection{Text-driven image generation}
Text-generated images is a method of converting text descriptions into images using artificial intelligence techniques. This technique has been implemented using Generative Adversarial Networks (GAN) or Convolutional Neural Networks (CNN). It can be used to generate visually realistic images, or to generate visually similar images. Diffusion models have recently been shown to generate high-quality synthetic images, especially when paired with bootstrapping techniques that provide a trade-off between diversity and fidelity.

The text-image generation task implemented by the team of Omri Avrahami \cite{avrahami2022blended} by using and combining a pre-trained language image model (CLIP) to direct editors to user-supplied text cues and using a denoising diffusion probability model (DDPM) to generate natural results.The team of Gwanghyun Kim \cite{kim2021diffusionclip} also used CLIP to propose DiffusionCLIP, a text-guided image processing method using a pre-trained diffusion model and CLIP loss.

Alex Nichol's \cite{nichol2021glide} team investigated a diffusion model for the text-conditional image synthesis problem and compared two different framing strategies:CLIP framing and classifier-free framing. We found that the latter is preferred by human evaluators in terms of photo-realism and caption similarity, and typically produces photo-realistic samples.

In the text conditional image generation task, Aditya Ramesh \cite{ramesh2022hierarchical} proposed a two-stage model:a priori generation of CLIP image embeddings for a given text title and decoder generation of images conditional on the image embeddings.

The team of Xihui Liu \cite{liu2023more} investigated fine-grained continuous control of this model class and introduced a new unified framework for semantic diffusion guidance that allows either linguistic or image guidance, or both. The bootstrap is injected into a pre-trained unconditional diffusion model using either image-text gradients or image matching scores, without retraining the diffusion model. The models also use CLIP guidance.

The team of Chitwan Saharia \cite{saharia2022photorealistic} proposed the Imagen model, a one text-to-image diffusion model with an unprecedented degree of realism and depth of language understanding.Imagen builds on a large transformer language model for understanding text and relies on the strength of the diffusion model in high-fidelity image generation.

Weilun Wang's \cite{wang2022semantic} team proposed a new DDPM-based framework for semantic image synthesis. Unlike previous conditional diffusion models that input semantic layouts and noisy images directly as inputs into the U-Net structure, which may not fully utilize the information in the input semantic mask, our framework handles semantic layouts and noisy images in a different way. It inputs the noisy image into the encoder of the U-Net structure and the semantic layout into the decoder by means of a multilayer spatial adaptive normalization operator.

Robin Rombach's \cite{rombach2022text} team proposed an approach based on retrieval enhanced diffusion models (rdms). In RDMs, a set of nearest neighbors is retrieved from an external database during the training process of each training instance, and the diffusion model is conditioned on these samples of information. A database with a specific style is then used in the sampling process to achieve images capable of generating a specific visual style.

The team of Shuyang Gu \cite{gu2022vector} proposed a vector quantization diffusion (VQ-Diffusion) model for text-to-image generation. The method is based on vector quantization variational self-encoder (VQ-VAE), whose potential space is modeled by a conditional variant of the recently developed denoising diffusion probability model (DDPM). The model is experimentally shown to be suitable for processing complex scenes and to substantially improve the quality of text-generated images.
\subsection{Text to Speech}
Text-to-speech is a natural language processing technique called Text-to-Speech Synthesis (TTS). It reads text aloud by converting that text into speech. TTS systems can use synthetic speech (generating speech by combining audio clips) or use recorded speech (generating speech by playing recorded speech). The denoising diffusion probability model (ddpm) has recently achieved leading performance in many generation tasks. However, the cost of the inherited iterative sampling process hinders their application in text-to-speech deployments. A group of scholars has faced the challenge of trying out diffusion models for text-to-speech synthesis tasks with some results.

Dongchao Yang's \cite{yang2022diffsound} team investigated the generation of sounds conditional on text cues and proposed a new text-to-speech generation framework consisting of a text encoder, a vector quantization variational autoencoder (VQ-V AE), a decoder, and a vocoder. The framework first converts the text features extracted by the text encoder into a speech spectrogram using a decoder with the help of VQ-VAE, and then converts the generated speech spectrogram into a waveform using a vocoder.

Heeseung Kim's \cite{kim2022guided} team proposed Guided-TTS, a high-quality text-to-speech (TTS) model using classifier bootstrapping that does not require any transcript of the target speaker. Guided-TTS combines an unconditional diffusion probability model and a separately trained phoneme classifier for classifier guidance.

The team of Rongjie Huang \cite{huang2022prodiff} proposed ProDiff, a progressive fast diffusion model for high-quality text-to-speech. Unlike previous work estimating data density gradients, ProDiff parameterizes the denoising model by directly predicting clean data to avoid the apparent quality degradation in accelerated sampling.

\subsection{Other Tasks}
\subsubsection{Text classification} generating stream and diffusion models are mainly trained on ordered data, Emiel Hoogeboom \cite{hoogeboom2021argmax} team presents in this paper two extensions of stream and diffusion for classified data (e.g. language or image segmentation):Argmax stream and multinomial diffusion. Where the flow is defined by a combination of a continuous distribution (e.g., normal flow) and an Argmax function.

\subsubsection{Semantic segmentation}: the team of Dmitry Baranchuk \cite{baranchuk2021label} demonstrated that diffusion models can also be used as a tool for semantic segmentation, especially when labeled data are scarce. The team studied intermediate activations from networks performing Markov steps of the reverse diffusion process. We show that these activations effectively capture semantic information from the input image and appear to be excellent pixel-level representations of the segmentation problem.

\subsubsection{Interpretable text modeling}: the team of Peiyu Yu \cite{yu2022latent} introduced a new symbiotic relationship between diffusion models and latent space ebm in a variational learning framework, called latent diffusion energy-based models. Combined with information bottlenecks, a regularization method based on geometric clustering was proposed to further improve the quality of learning latent spaces. The model is used in interpretable text modeling.

\section{Visualization Analysis}
In the follow-up study, the cited diffusion model literature is summarized and some useful information is obtained in this paper.

In terms of literature publication years, the earliest three papers on diffusion models for NLP were published in 2021 and 19 in 2022. In contrast, the DDPM model was proposed in 2020, which means that the research on the application of diffusion models to the field of NLP started around the same time and achieved some results in 2021 and a breakthrough in 2022, which also indicates the rapid development of diffusion models.

Among the publications, arXiv has the largest number of papers, followed by CVPR, and ACM-MM has the smallest number. arXiv is an important publication in the field of deep learning, ICLR, ICML, and NeurIPS, ACM-MM is an important publication in the field of computer multimodality, and CVPR is an important publication involving computer vision and pattern recognition. arXiv is mostly research results that have not been formally published. The content of Figure 6 illustrates the importance of the deep learning field for applying diffusion models to the NLP domain.

In terms of citations in the literature, divided by application scenarios, the mean citation in the field of text generation is about 15, with generally low citations; the mean citation in the field of text-to-image generation is about 147, with a maximum of 563 and a minimum citation; and the citations in the field of text-to-speech are all below 10. This indicates to some extent that the main research direction of diffusion models in the NLP field is to use them in text conditional image generation tasks.

\section{Conclusion}
This paper reviews the progress made by the research community in developing and applying diffusion models to natural language processing tasks, and identifies the main formula for diffusion modeling based on DDPMs. In addition, this paper also collects and analyzes the relevant information of the cited literature. According to the literature review results and related materials, the current main research directions of diffusion models in NLP field are text generation and text-conditioned image generation, and some scholars have also started to conduct research on other aspects such as text-to-speech, semantic segmentation, explainable text modeling, etc.

% or


\bibliographystyle{IEEEtran}
\bibliography{ref}
% \appendices

% % you can choose not to have a title for an appendix
% % if you want by leaving the argument blank
% \section{}

% % use section* for acknowledgment
% \section*{Acknowledgment}


\ifCLASSOPTIONcaptionsoff
  \newpage
\fi

\end{document}